\documentclass[acmtog, authorversion,nonacm]{acmart}

\AtBeginDocument{%
  \providecommand\BibTeX{{%
    \normalfont B\kern-0.5em{\scshape i\kern-0.25em b}\kern-0.8em\TeX}}}

\acmJournal{TALLIP}
\acmVolume{}
\acmNumber{}
\acmArticle{}
\acmMonth{}

\usepackage{enumitem}
\usepackage{graphicx}
\usepackage{tabularx}
\usepackage{subfig}
\usepackage{booktabs, multirow} 
\usepackage{soul}

\begin{document}

\title{L3Cube-IndicSBERT: A simple approach for learning cross-lingual sentence representations using multilingual BERT
}

\author{Samruddhi Deode}
\authornote{Authors contributed equally to this research.}
\email{samruddhi321@gmail.com}
\orcid{0009-0008-6559-2752}
\affiliation{%
  \institution{MKSSS' Cummins College of Engineering for Women; }
  \institution{L3Cube Pune}
  \city{Pune}
  \state{Maharashtra}
  \country{India}
}

\author{Janhavi Gadre}
\authornotemark[1]
\email{janhavi.gadre@gmail.com}
\orcid{0009-0007-7967-396X}
\affiliation{%
  \institution{MKSSS' Cummins College of Engineering for Women;}
  \institution{L3Cube Pune}
  \city{Pune}
  \state{Maharashtra}
  \country{India}
}

\author{Aditi Kajale}
\authornotemark[1]
\email{aditi1.y.kajale@gmail.com}
\orcid{0009-0009-8170-0925}
\affiliation{%
  \institution{MKSSS' Cummins College of Engineering for Women;}
  \institution{L3Cube Pune}
  \city{Pune}
  \state{Maharashtra}
  \country{India}
}

\author{Ananya Joshi}
\authornotemark[1]
\email{joshiananya20@gmail.com}
\orcid{0009-0008-9015-0550}
\affiliation{%
  \institution{MKSSS' Cummins College of Engineering for Women;}
  \institution{L3Cube Pune}
  \city{Pune}
  \state{Maharashtra}
  \country{India}
}

\author{Raviraj Joshi}
\email{ravirajoshi@gmail.com}
\orcid{0000-0003-1892-1812}
\affiliation{%
  \institution{Indian Institute of Technology Madras;}
  \institution{L3Cube Pune}
  \city{Chennai}
  \state{Tamil Nadu}
  \country{India}
}

\renewcommand{\shortauthors}{S. Deode et al.}

\begin{abstract}
The multilingual Sentence-BERT (SBERT) models map different languages to common representation space and are useful for cross-language similarity and mining tasks.  
We propose a simple yet effective approach to convert vanilla multilingual BERT models into multilingual sentence BERT models using synthetic corpus. We simply aggregate translated NLI or STS datasets of the low-resource target languages together and perform SBERT-like fine-tuning of the vanilla multilingual BERT model. We show that multilingual BERT models are inherent cross-lingual learners and this simple baseline fine-tuning approach without explicit cross-lingual training yields exceptional cross-lingual properties. We show the efficacy of our approach on 10 major Indic languages and also show the applicability of our approach to non-Indic languages German and French.
Using this approach, we further present L3Cube-IndicSBERT, the first multilingual sentence representation model specifically for Indian languages Hindi, Marathi, Kannada, Telugu, Malayalam, Tamil, Gujarati, Odia, Bengali, and Punjabi. The IndicSBERT exhibits strong cross-lingual capabilities and performs significantly better than alternatives like LaBSE, LASER, and paraphrase-multilingual-mpnet-base-v2 on Indic cross-lingual and monolingual sentence similarity tasks. We also release monolingual SBERT models for each of the languages and show that IndicSBERT performs competitively with its monolingual counterparts. These models have been evaluated using embedding similarity scores and classification accuracy. 
\end{abstract}



\maketitle

\section{Introduction}
Natural Language Processing (NLP) is an interdisciplinary field that focuses on developing techniques to process and understand human language \cite{otter2020survey}. Semantic Textual Similarity (STS) is a crucial task in NLP, which measures the equivalence between the meaning of two or more text segments ~\cite{agirre2013sem,cer2017semeval}. The aim of STS is to identify the semantic similarity between text inputs, taking into account their meaning rather than just surface features like word frequency and length \cite{adifine}. The concept is widely used in various NLP applications, including question-answering \cite{huang2020recent}, information retrieval \cite{li2016deep}, text generation \cite{iqbal2022survey}, etc.

One common tool used for this purpose is BERT (Bidirectional Encoder Representations from Transformers) ~\cite{devlin2019bert}, a pre-trained transformer-based language model that has achieved state-of-the-art performance on a wide range of NLP tasks. However, BERT is not well-suited for semantic similarity tasks as it is trained to predict masked words in a sentence, which does not directly optimize for semantic similarity \cite{zhang2020semantics}. To address this limitation, Sentence-BERT (SBERT) ~\cite{reimers2019sentence} was proposed, a modified version of the BERT architecture designed to generate sentence representations for the improved semantic similarity between sentences. The SBERT makes use of a siamese network \cite{koch2015siamese} and is trained using specific datasets like STS, resulting in representations specifically geared for semantic similarity. 

Recent works are focused on multilingual SBERT models capable of encoding sentences from different languages to the same representation space ~\cite{schwenk2017learning, yang2020multilingual}. With these models, it is possible to extend NLP tasks to different languages without training a language-specific model. These multilingual models often employ teacher-student training ~\cite{heffernan2022bitext,reimers2020making} or are based on translation ranking tasks ~\cite{feng2022languageagnostic}. These methods make use of parallel translation corpus in target languages for training a cross-lingual model ~\cite{tan2022bitext,Artetxe_2019,conneau2018xnli}. Even vanilla multilingual BERT models have been shown to have surprisingly good cross-lingual properties \cite{pires2019multilingual,wu2020all}. However, their performance is not good as the multilingual sentence BERT models.

\begin{figure}[]
\centering
\includegraphics[scale=0.5]{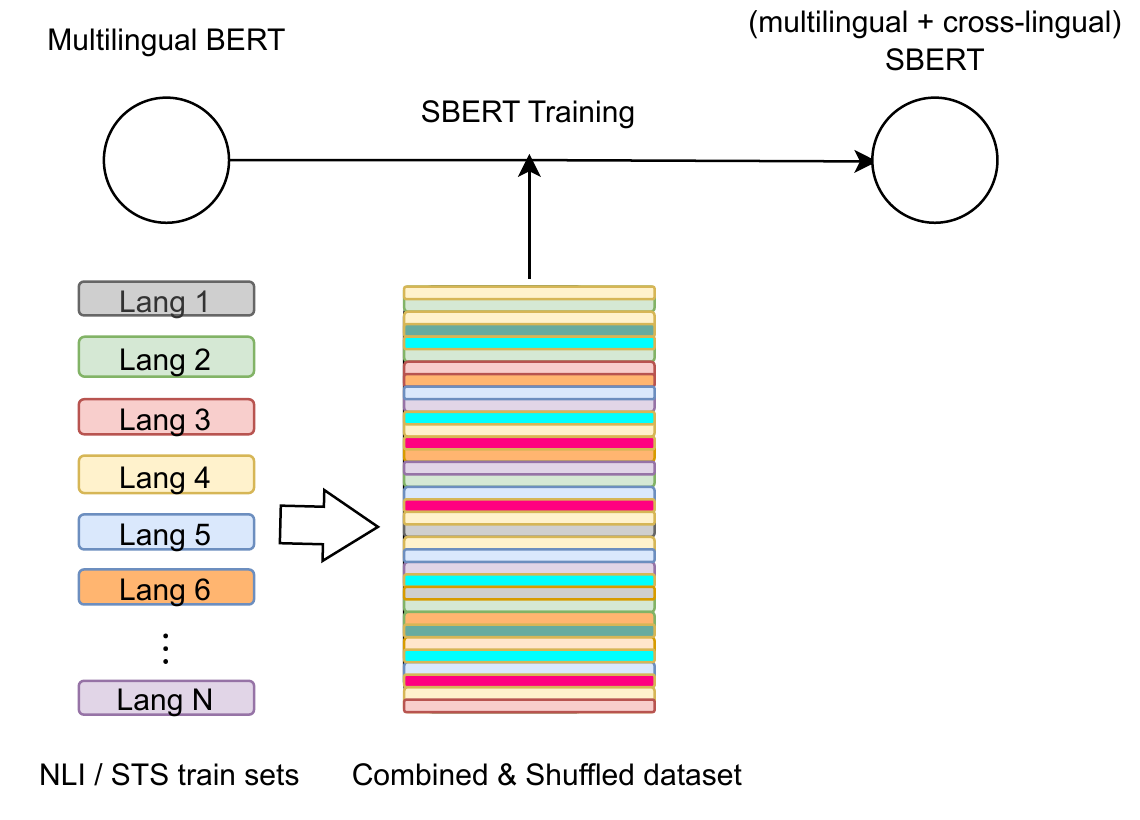}
\caption{An embarrassingly simple approach for learning cross-lingual sentence representations using synthetic monolingual corpus}\label{fig:0}
\end{figure}

In this work, we propose a simple approach to learning cross-lingual sentence representations without using any parallel corpus. We leverage pre-trained multilingual transformer models and fine-tune them using our mixed training strategy, as depicted in Figure \ref{fig:0}. We mix the monolingual translated NLI / STSb corpus for target languages and fine-tune the multilingual BERT model in an SBERT setup. We show that this simple mixed data training is sufficient to train a multilingual SBERT model with strong cross-lingual properties. This strategy is capable of significantly amplifying the cross-lingual properties of the existing vanilla multilingual BERT model. Our approach is inspired by a recent work ~\cite{joshi2022l3cubemahasbert} that shows that translated STSb and NLI can be used to train high-quality monolingual SBERT models. 

We present L3Cube-IndicSBERT a multilingual SBERT model for 10 Indian regional languages Hindi, Marathi, Kannada, Telugu, Malayalam, Tamil, Gujarati, Odia, Bengali, Punjabi, and English. The IndicSBERT uses MuRIL ~\cite{khanuja2021muril} as the base model and performs better than existing multilingual/cross-lingual models like LASER, LaBSE, and paraphrase-multilingual-mpnet-base-v2. These models are compared on monolingual and cross-lingual sentence similarity tasks. We also evaluate these models on real text classification datasets to show that the synthetic data training generalizes well to real datasets. Further, we also release monolingual SBERT models for individual languages to show that IndicSBERT performs competitively with the monolingual variants.
\\\\
Our main contributions are as follows:
\begin{itemize}
\item[$\bullet$]We propose a simple strategy to train cross-lingual sentence representations using a pre-trained multilingual BERT model and synthetic NLI/STS data. Unlike previous approaches, it does not use any cross-lingual data or any complex training strategy.
\item[$\bullet$]We present \textbf{IndicSBERT}\footnote[1]{\url{https://huggingface.co/l3cube-pune/indic-sentence-bert-nli}}\footnote[2]{\url{https://huggingface.co/l3cube-pune/indic-sentence-similarity-sbert}}, the first multilingual SBERT model trained specifically for Indic languages. The model performs better than state-of-the-art LaBSE and paraphrase-multilingual-mpnet-base-v2 models.
\item[$\bullet$]We also release monolingual SBERT models for 10 Indic languages. To the best of our knowledge, this work is first to introduce the majority of these models.
\end{itemize}
The subsequent sections of the paper are organized as follows: \\Section 2 examines prior research on improving BERT performance and surveys previous work on sentence-BERT models. In Section 3.1, the datasets utilized in this study are outlined, while Section 3.2 provides details on the various models used and Section 3.3 delves into the specifics of the SBERT training procedure. Section 4 outlines the evaluation strategy used for the models and presents the key findings from our experiment. Finally, the paper concludes with a summary of all observations. This work is released as a part of the MahaNLP project.

\section{Related Work}
BERT~\cite{devlin2019bert} (Bidirectional Encoder Representations from Transformers) is a pre-trained transformer network that is widely regarded as one of the best language models for natural language processing (NLP) tasks, such as text categorization and named entity recognition. Vanilla BERT models serve as a starting point for many NLP tasks, and researchers and practitioners often use their pre-trained weights to fine-tune models on specific tasks.

mBERT~\cite{devlin2019bert} (Multilingual BERT) is a BERT-based language model, pretrained using MLM (Masked Language Modeling) objective on 104 different languages. XLM-RoBERTa ~\cite{conneau2020unsupervised} is another large-scale, cross-lingual language model developed by Facebook, trained on 100 different languages, making it a highly effective model for multilingual NLP tasks. A study~\cite{rust2021good} reveals that pretraining data size and a designated monolingual tokenizer are important factors that affect performance, and replacing the original multilingual tokenizer with a specialized monolingual tokenizer improves the downstream performance of the multilingual model for most languages and tasks. Despite numerous attempts at building better and bigger multilingual language models (MLLMs), as shown in ~\cite{doddapaneni2021primer}, there has been limited research focused on creating models specifically for low-resource languages. In ~\cite{pfeiffer2021unks}, the authors show the effectiveness of novel data-efficient methods using matrix factorization and lexically overlapping tokens for the adaptation of pre-trained multilingual models to low-resource languages and unseen scripts.

For the Indian languages, the available multilingual models include IndicBERT ~\cite{kakwani2020indicnlpsuite}. It follows the architecture of the original BERT model but is trained on a large corpus of text from several Indian languages. Another multilingual model, MuRIL ~\cite{khanuja2021muril} (Multilingual Representations for Indian Languages) has been pre-trained on 17 Indic languages. 

Sentence embedding models \cite{cer2018universal,conneau2017supervised,yang2020multilingual,ni2022sentence} are superior to word embedding models \cite{pennington2014glove,bojanowski2017enriching,peters-etal-2018-deep,ethayarajh2019contextual} as they capture the meaning of the entire sentence rather than individual words. While BERT is trained to generate word embeddings, Sentence-BERT~\cite{reimers2019sentence} modifies the architecture and fine-tunes the pre-trained BERT model for generating sentence embeddings. SBERT also includes additional training methods, such as the Siamese and triplet network architectures, that allow for more effective training of sentence embeddings. The SBERT model is trained using supervised datasets like NLI and STS that help in understanding the sentence semantics.

Numerous unsupervised methods have been proposed that learn meaningful sentence embeddings directly from text without the need for labeled training data. These include TSDAE  which rebuilds noisy versions of input data while maintaining the data's original semantics. SimCSE is a contrastive learning technique that learns to encode the semantic similarity of phrase pairs into their embeddings. However, in this work, we focus solely on the supervised approaches for learning sentence embeddings.

LaBSE~\cite{feng2022languageagnostic}, a sentence-BERT model is designed to generate language-agnostic sentence embeddings that can be used for cross-lingual NLP tasks, while LASER~\cite{Artetxe_2019} is a multilingual sentence embedding model that generates high-quality sentence embeddings for multiple low-resource languages. These models have been explicitly trained using parallel translation corpus. Similarly, by aligning the embeddings of parallel sentences in many languages, Cross-Lingual Transfer (CT)~\cite{Artetxe_2019} technique learns a shared space for sentence embeddings across multiple languages. Thus, in the multilingual category, several BERT, as well as Sentence-BERT models, have been developed to date.

However, monolingual models are typically found to be performing better than multilingual ones. In a previous study ~\cite{scheible2020gottbert}, a German RoBERTa-based BERT model, with slight adjustments to its hyperparameters, was found to yield superior results than all other German and multilingual BERT models. Similarly, in ~\cite{Straka_2021} a Czech RoBERTa language model has been shown to perform better than other Czech and multilingual models. In ~\cite{Velankar_2022} and ~\cite{joshi2022l3cube}, monolingual BERT models for the Marathi language were studied and found to perform better than their multilingual counterparts. Hence, to obtain an improved performance with rich sentence embeddings, monolingual Sentence-BERT models were proposed. Similarly, in this study, we propose monolingual SBERT models for the ten most prominent Indic languages. Additionally, we also propose a multilingual model tailored specifically to these languages. Considering that other multilingual models are trained to support a greater number of languages, our model is better suited for Indian languages, as it is specifically optimized for them.

\section{Experimental Setup}
\subsection{Datasets}
The results shown in ~\cite{joshi2022l3cubemahasbert}, indicate the efficacy of using synthetic datasets in creating MahaSBERT-STS and HindSBERT-STS. Thus, we utilize the machine-translated IndicXNLI and STSb datasets for training our models. Our models are evaluated on the synthetic STSb dataset, as well as on real-world classification datasets. The 3 datasets are described below.
\\

The \textbf{IndicXNLI}\footnote[3]{\url{https://github.com/divyanshuaggarwal/IndicXNLI}}  dataset comprises of English XNLI data translated into eleven Indian languages including Hindi and Marathi. To train the monolingual Sentence-BERT models, we use the training samples of the corresponding language from IndicXNLI. To ensure balanced training data for the multilingual IndicSBERT, we combine and randomly shuffle the IndicXNLI training samples of ten languages.  

The \textbf{STS benchmark (STSb)\footnote[4]{\url{https://huggingface.co/datasets/stsb\_multi\_mt}}} dataset is commonly utilized for evaluating supervised Semantic Textual Similarity (STS) systems. The dataset includes 8628 sentence pairs from captions, news, and forums and is divided into 5749 for training, 1500 for development and 1379 for testing. To make the dataset accessible for all ten Indian languages used in this study, we translate it using Google Translate and use the resulting train samples of the corresponding language for each monolingual model and a combined dataset of ten languages for the multilingual model. We use the testing samples from the corresponding translated STSb dataset to evaluate each model based on the embedding similarity metric. For evaluating the cross-lingual property, we construct a dataset of STSb sentence pairs with each pair comprising two sentences from different languages.

We also evaluate the models on real text classification datasets. We perform this evaluation to show that the sentence representations from the models trained using synthetic datasets also generalize well to real datasets. We choose the \textbf{IndicNLP news article classification datasets}~\cite{kunchukuttan2020ai4bharatindicnlp} for the purpose of evaluation. The classification datasets consist of train, validation, and test sets in an 8:1:1 ratio.

We also apply a series of preprocessing steps, such as removing punctuation, URLs, hashtags, Roman characters and blank spaces, to ensure that the data is suitable for our experiments.

\subsection{Models}
BERT is a deep, bi-directional model based on the Transformer architecture, which has been trained on a large, unlabeled corpus. Multiple pre-trained BERT models, both monolingual and multilingual, are publicly available for use. In our experiment, we use different BERT models, including both monolingual and multilingual ones which are described below. The training procedure is applied over some of these models which serve as a base for creating Sentence-BERT. 

\subsubsection{\textbf{Multilingual BERT models:}}

\begin{itemize}
  \item[$\bullet$] \textbf{mBERT\footnote[5]{\url{https://huggingface.co/bert-base-multilingual-cased}}}: A pre-trained multilingual BERT-base model that has been trained on 104 languages using a combination of the next sentence prediction (NSP) and Masked Language Modeling (MLM) ~\cite{devlin2019bert} objectives.\\

  \item[$\bullet$] \textbf{MuRIL\footnote[6]{\url{https://huggingface.co/google/muril-base-cased}} (Multilingual Representations for Indian Languages)}: a BERT-based model which supports 17 Indian languages ~\cite{khanuja2021muril}. It is pre-trained using masked language modeling and next-sentence prediction objectives on parallel data, which includes the translations as well as transliterations on each of the 17 monolingual corpora. It has shown state-of-the-art performance on a variety of language understanding tasks.\\

  \item[$\bullet$] \textbf{LaBSE\footnote[7]{\url{https://huggingface.co/setu4993/LaBSE} } ~\cite{feng2022languageagnostic} (Language-agnostic BERT sentence embedding)}: It is a transformer-based model that learns language-agnostic sentence representations through a cross-lingual sentence retrieval task. It was trained on parallel sentence pairs from 109 languages using a Siamese network based on the BERT architecture. The model's ability to support 109 languages makes it a powerful tool for multilingual applications and cross-lingual natural language processing tasks.\\

 \item[$\bullet$] \textbf{paraphrase-multilingual-mpnet-base-v2\footnote[8]{\url{https://huggingface.co/sentence-transformers/paraphrase-multilingual-mpnet-base-v2}}}:  It is based on a Multilingual Pretrained Transformer (MPT) architecture. This model supports 50 languages and is trained on a paraphrase identification task. It has achieved state-of-the-art performance on the paraphrase identification task on several benchmark datasets.\\

 \item[$\bullet$] \textbf{LASER~\cite{Artetxe_2019}} (Language-Agnostic SEntence Representations): This model from Facebook is trained on large parallel corpora for cross-lingual language understanding (XLU) task. It uses a multilingual encoder-decoder architecture, where the encoder is a five-layer bidirectional LSTM. It can generate superior-quality cross-lingual sentence embeddings for over 90 languages and outperforms other models on cross-lingual tasks including machine translation, sentiment analysis, and cross-lingual information retrieval.
   
 \end{itemize}

 \subsubsection{\textbf{Monolingual BERT models:}\\}
We also use the monolingual BERT models for the 10 Indic languages, released by L3cube-Pune\footnote[9]{\url{ https://huggingface.co/l3cube-pune}} as the base models. These models are termed as HindBERT, MahaBERT \cite{joshi2022l3cube}, KannadaBERT, TeluguBERT, MalayalamBERT, TamilBERT, GujaratiBERT, OdiaBERT, BengaliBERT, and PunjabiBERT. Further details about these models can be found in ~\cite{joshi2023l3cubehindbert}.
\\

\begin{figure}[]
\centering
\includegraphics[scale=0.45]{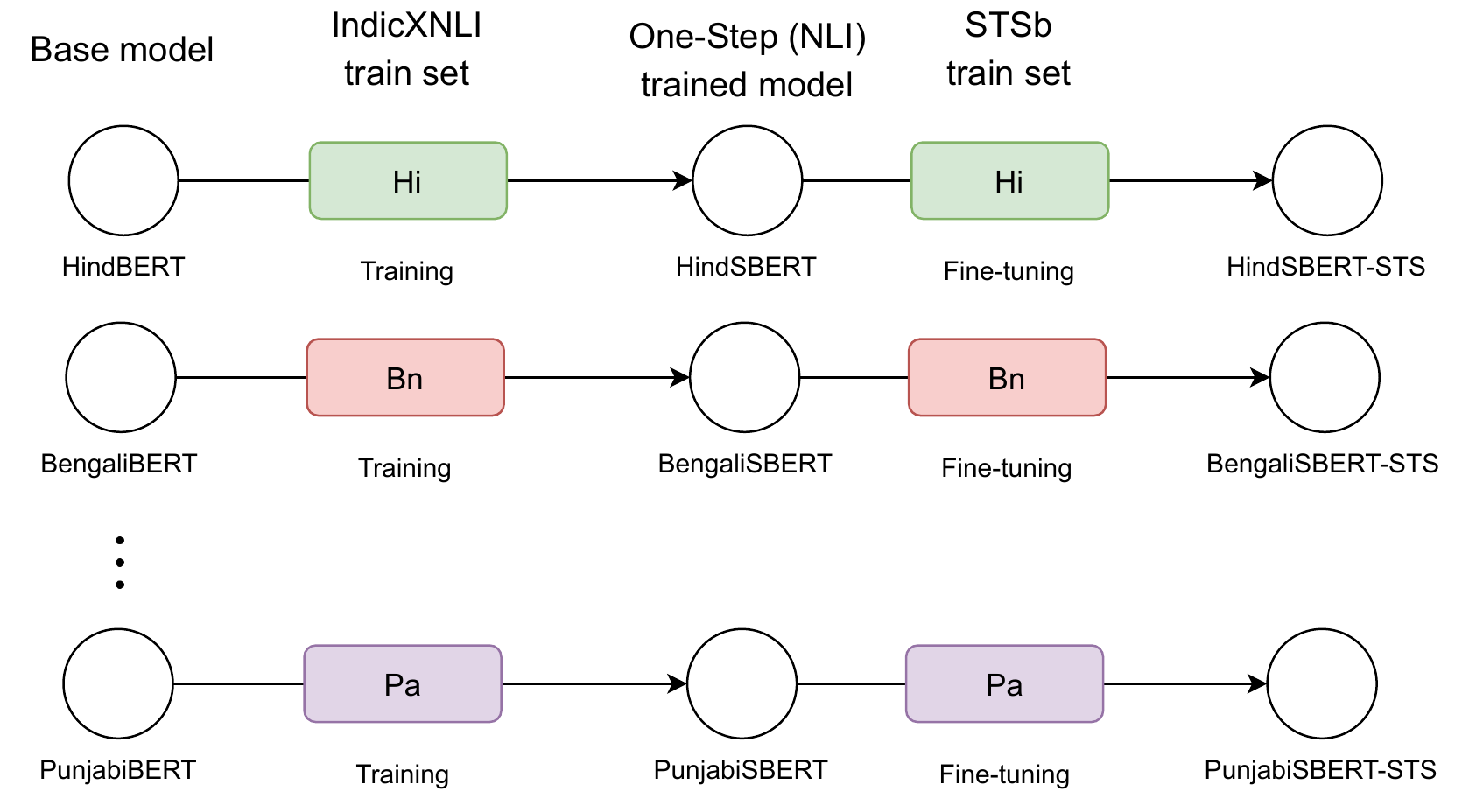}
\caption{Two-step (NLI + STS) training of the monolingual SBERT models}\label{fig:1}
\end{figure}
\begin{figure}[]
\centering
\includegraphics[scale=0.5]{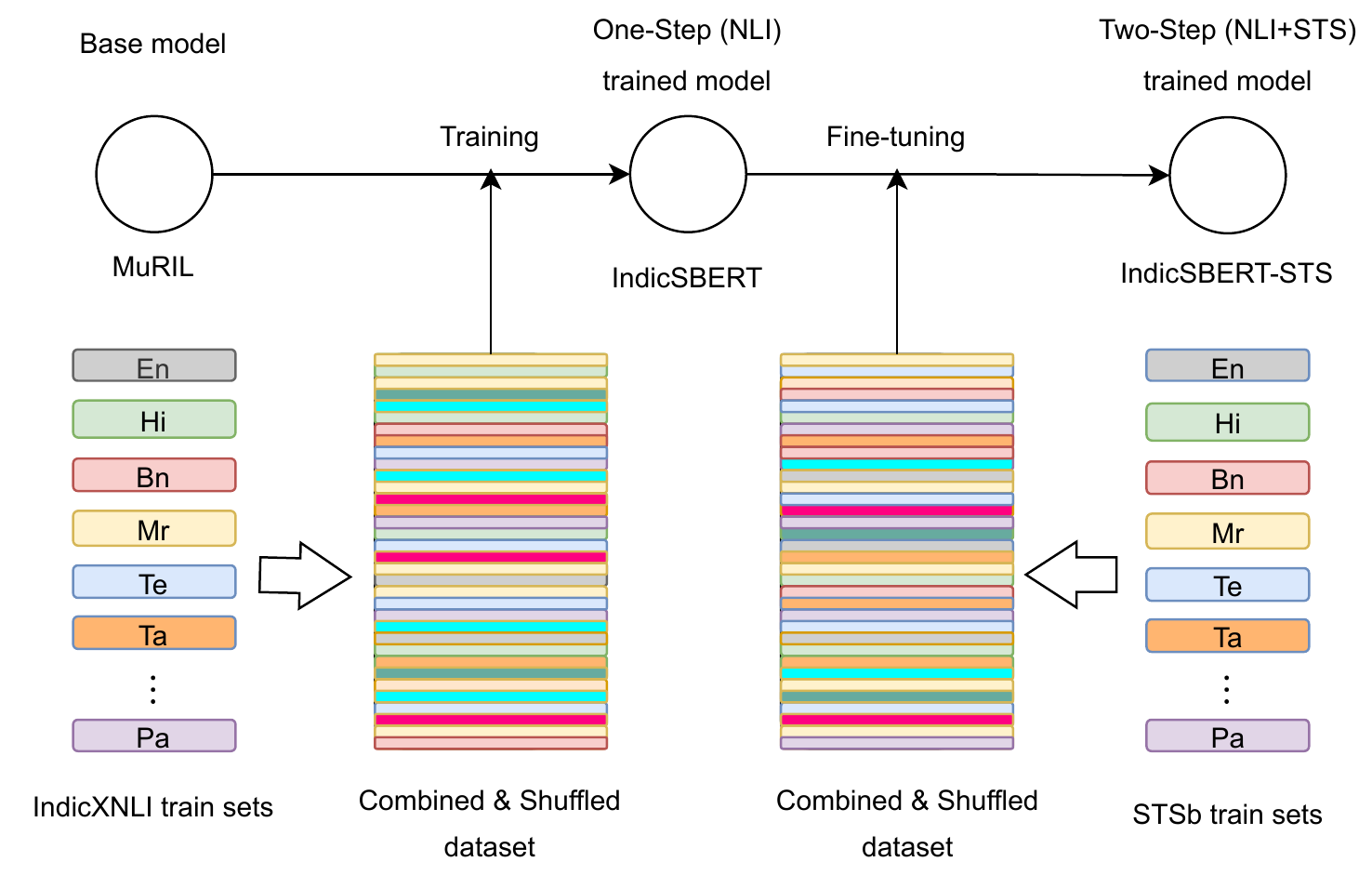}
\caption{Two-step (NLI + STS) training of the multilingual IndicSBERT models}\label{fig:2}
\end{figure}

\subsection{SBERT Training}
In order to achieve competitive performance, sentence embedding models typically require significant amounts of training data and fine-tuning over the target task. Unfortunately, in many scenarios, only limited amounts of training data are available. Several unsupervised and semi-supervised approaches have been proposed to overcome the lack of a large training dataset. However, the models trained using unsupervised techniques give inferior performance than those trained using supervised learning.

In this work, we, therefore, use a supervised training approach, wherein we address the scarcity of specialized datasets, such as NLI and STS, in Indian languages by machine translating the English versions of these datasets into the respective Indian languages. We follow a two-step procedure to train the monolingual SBERT models and the  multilingual IndicSBERT model. The monolingual BERT model serves as the base for monolingual SBERT while MuRIL serves as the base model for IndicSBERT.
\\\\ 
In the \textbf{first step} of the training procedure, natural language inference, or textual entailment, is performed. This task involves determining the logical relationship between a premise and hypothesis, represented as text sequences. The aim is to classify the relationship into three categories: entailment (hypothesis can be inferred from the premise), contradiction (negation of the hypothesis can be inferred from the premise), or neutral (no clear relationship between the two). In this step, the base model is trained on the IndicXNLI dataset, which consists of 392702 sentence pairs, each labeled as entailment, contradiction, or neutral.

To improve the effectiveness of the model, we utilize the Multiple Negatives Ranking Loss function instead of the Softmax-Classification-Loss used in ~\cite{reimers2019sentence}. This is because the Multiple Negatives Ranking Loss, which considers multiple negative samples simultaneously, is better suited for similarity-based problems where the goal is to learn similarities and dissimilarities between examples. This results in a more complex decision boundary and improves the model's robustness to outliers and variations in data.

The training data consists of triplets: [(a1, b1, c1), …, (an, bn, cn)], where (ai, bi) are considered similar sentences and (ai, ci) are dissimilar. An entry for bi is randomly picked from the set of sentences labeled as 'entailment' for ai, and an entry for ci is picked from the set of sentences labeled as 'contradiction' for ai, referred to as hard-negatives. Although they are similar to ai and bi on a lexical level, they mean different things on a semantic level. The model is trained using 1 epoch, with a batch size of 4, AdamW optimizer, and a learning rate of 2e-05. The AdamW optimizer extends the Adam optimizer and adds weight decay regularization to prevent overfitting and improve the model's generalization.

The models obtained after applying the first step (NLI only) of training are named as \textbf{MahaSBERT\footnote[10]{\url{ https://huggingface.co/l3cube-pune/marathi-sentence-bert-nli}}, HindSBERT\footnote[11]{\url{https://huggingface.co/l3cube-pune/hindi-sentence-bert-nli}}, KannadaSBERT\footnote[12]{\url{https://huggingface.co/l3cube-pune/kannada-sentence-bert-nli}}, TeluguSBERT\footnote[13]{\url{https://huggingface.co/l3cube-pune/telugu-sentence-bert-nli}}, MalayalamSBERT\footnote[14]{\url{https://huggingface.co/l3cube-pune/malayalam-sentence-bert-nli}}, TamilSBERT\footnote[15]{\url{ https://huggingface.co/l3cube-pune/tamil-sentence-bert-nli}}, GujaratiSBERT\footnote[16]{\url{ https://huggingface.co/l3cube-pune/gujarati-sentence-bert-nli}}, OdiaSBERT\footnote[17]{\url{https://huggingface.co/l3cube-pune/odia-sentence-bert-nli}}, BengaliSBERT\footnote[18]{\url{ https://huggingface.co/l3cube-pune/bengali-sentence-bert-nli}}, and PunjabiSBERT\footnote[19]{\url{ https://huggingface.co/l3cube-pune/punjabi-sentence-bert-nli}}} that are made publicly available.
\\\\
In the \textbf{second step}, the model from step one is fine-tuned using the translated STSb dataset. The STS benchmark is a commonly used dataset for evaluating the performance of NLP models in determining the similarity between two pieces of text. It comprises sentence pairs with human-annotated similarity scores on a scale of 0-5. The fine-tuning process uses the Cosine Similarity Loss as the loss function, which measures the similarity between two vectors in a multi-dimensional space. Cosine similarity loss considers the angle between vectors rather than their magnitudes, making it a robust measure of similarity. It is derived by computing the vectors for the two input texts, taking the dot product of the two vectors and dividing it by the product of the magnitudes of the two vectors. The result is a value between -1 and 1, where -1 indicates complete dissimilarity and 1 indicates complete similarity. The training process involves 4 epochs with Cosine Similarity Loss as the loss function and uses an AdamW optimizer with a learning rate of 2e-05.
\\\\
The final models obtained after applying the two-step procedure (NLI + STS) are named as \textbf{MahaSBERT-STS\footnote[20]{\url{https://huggingface.co/l3cube-pune/marathi-sentence-similarity-sbert}}, HindSBERT-STS\footnote[21]{\url{https://huggingface.co/l3cube-pune/hindi-sentence-similarity-sbert}}, KannadaSBERT-STS\footnote[22]{\url{https://huggingface.co/l3cube-pune/kannada-sentence-similarity-sbert}}, TeluguSBERT-STS\footnote[23]{\url{https://huggingface.co/l3cube-pune/telugu-sentence-similarity-sbert}}, MalayalamSBERT-STS\footnote[24]{\url{https://huggingface.co/l3cube-pune/malayalam-sentence-similarity-sbert}}, TamilSBERT-STS\footnote[25]{\url{ https://huggingface.co/l3cube-pune/tamil-sentence-similarity-sbert}}, GujaratiSBERT-STS\footnote[26]{\url{ https://huggingface.co/l3cube-pune/gujarati-sentence-similarity-sbert}}, OdiaSBERT-STS\footnote[27]{\url{https://huggingface.co/l3cube-pune/odia-sentence-similarity-sbert}}, BengaliSBERT-STS\footnote[28]{\url{ https://huggingface.co/l3cube-pune/bengali-sentence-similarity-sbert}}, and PunjabiSBERT-STS\footnote[29]{\url{ https://huggingface.co/l3cube-pune/punjabi-sentence-similarity-sbert}}} and are made publicly available. In addition to the models mentioned above, we also release the multilingual SBERT models named \textbf{IndicSBERT} and \textbf{IndicSBERT-STS}. These multilingual models support the 11 languages of English, Hindi, Marathi, Kannada, Telugu, Malayalam, Tamil, Gujarati, Odia, Bengali, and Punjabi.

\renewcommand*{\thefootnote}{\fnsymbol{footnote}}

\begin{table}[!htp]\centering
\caption{Embedding similarity scores of monolingual BERT and SBERT models}\label{tab:1}
\scriptsize
\resizebox{\columnwidth}{!}{
\begin{tabular}{|l|ccc|ccc|ccc|ccc|}\toprule
&\multicolumn{9}{c|}{\textbf{Multilingual base}} &\multicolumn{3}{c|}{\textbf{Monolingual base}} \\\midrule
\textbf{Base model:} &\multicolumn{3}{c|}{\textbf{mBERT}} &\multicolumn{3}{c|}{\textbf{MuRIL}} &\multicolumn{3}{c|}{\textbf{LaBSE}} &\multicolumn{3}{c|}{\textbf{BERT}} \\\midrule
\textbf{Training steps\footnotemark[8]} &\textbf{0} &\textbf{1} &\textbf{2} &\textbf{0} &\textbf{1} &\textbf{2} &\textbf{0} &\textbf{1} &\textbf{2} &\textbf{0} &\textbf{1} &\textbf{2} \\\midrule
\textbf{Hindi (hi)} &0.49 &0.64 &\textbf{0.75} &0.52 &0.74 &\textbf{0.83} &0.72 &0.75 &\textbf{0.84} &0.5 &0.77 &\textbf{0.85} \\
\textbf{Bengali (bn)} &0.5 &0.65 &\textbf{0.75} &0.55 &0.74 &\textbf{0.82} &0.71 &0.75 &\textbf{0.81} &0.5 &0.72 &\textbf{0.81} \\
\textbf{Marathi (mr)} &0.47 &0.65 &\textbf{0.72} &0.56 &0.74 &\textbf{0.81} &0.7 &0.75 &\textbf{0.82} &0.54 &0.77 &\textbf{0.83} \\
\textbf{Telugu (te)} &0.53 &0.62 &\textbf{0.73} &0.6 &0.71 &\textbf{0.8} &0.73 &0.73 &\textbf{0.81} &0.58 &0.72 &\textbf{0.8} \\
\textbf{Tamil (ta)} &0.49 &0.65 &\textbf{0.75} &0.6 &0.72 &\textbf{0.8} &0.72 &0.74 &\textbf{0.82} &0.59 &0.72 &\textbf{0.8} \\
\textbf{Gujarati (gu)} &0.47 &0.65 &\textbf{0.74} &0.58 &0.72 &\textbf{0.8} &0.73 &0.73 &\textbf{0.82} &0.55 &0.74 &\textbf{0.82} \\
\textbf{Kannada (kn)} &0.52 &0.68 &\textbf{0.75} &0.6 &0.75 &\textbf{0.82} &0.72 &0.76 &\textbf{0.82} &0.57 &0.74 &\textbf{0.82} \\
\textbf{Odia (or)\footnotemark[2]} &- &- &\textbf{-} &0.45 &0.58 &\textbf{0.69} &0.6 &0.6 &\textbf{0.73} &0.45 &0.59 &\textbf{0.71} \\
\textbf{Malayalam (ml)} &0.46 &0.57 &\textbf{0.67} &0.53 &0.66 &\textbf{0.74} &0.66 &0.66 &\textbf{0.74} &0.5 &0.69 &\textbf{0.76} \\
\textbf{Punjabi (pa)} &0.43 &0.59 &\textbf{0.68} &0.45 &0.65 &\textbf{0.74} &0.64 &0.67 &\textbf{0.75} &0.5 &0.68 &\textbf{0.75} \\
\bottomrule
\end{tabular}}
\end{table}

\begin{table}[!htp]\centering
\caption{Classification accuracy of monolingual BERT and SBERT models}\label{tab:2}
\scriptsize
\resizebox{\columnwidth}{!}{
\begin{tabular}{|l|ccc|ccc|ccc|ccc|}\toprule
&\multicolumn{9}{c|}{\textbf{Multilingual base}} &\multicolumn{3}{c|}{\textbf{Monolingual base}} \\\midrule
\textbf{Base model:} &\multicolumn{3}{c|}{\textbf{mBERT}} &\multicolumn{3}{c|}{\textbf{MuRIL}} &\multicolumn{3}{c|}{\textbf{LaBSE}} &\multicolumn{3}{c|}{\textbf{BERT}} \\\midrule
\textbf{Training steps\footnotemark[8]} &\textbf{0} &\textbf{1} &\textbf{2} &\textbf{0} &\textbf{1} &\textbf{2} &\textbf{0} &\textbf{1} &\textbf{2} &\textbf{0} &\textbf{1} &\textbf{2} \\\midrule
\textbf{Hindi (hi)} &0.62 &0.6 &0.62 &0.67 &0.7 &0.69 &0.68 &0.64 &0.65 &0.7 &0.69 &0.68 \\
\textbf{Bengali (bn)} &0.97 &0.96 &0.97 &0.97 &0.98 &0.98 &0.98 &0.98 &0.97 &0.98 &0.98 &0.98 \\
\textbf{Marathi (mr)} &0.98 &0.97 &0.97 &0.97 &0.98 &0.98 &0.98 &0.99 &0.99 &0.98 &0.98 &0.99 \\
\textbf{Telugu (te)} &0.98 &0.97 &0.97 &0.99 &0.99 &0.99 &0.99 &0.99 &0.99 &0.99 &0.95 &0.99 \\
\textbf{Tamil (ta)} &0.96 &0.96 &0.95 &0.96 &0.97 &0.97 &0.96 &0.97 &0.96 &0.96 &0.97 &0.97 \\
\textbf{Gujarati (gu)} &0.95 &0.94 &0.93 &0.97 &0.98 &0.99 &0.99 &0.96 &0.99 &0.98 &0.99 &0.99 \\
\textbf{Kannada (kn)} &0.96 &0.95 &0.94 &0.97 &0.97 &0.97 &0.96 &0.96 &0.97 &0.97 &0.97 &0.97 \\
\textbf{Odia (or)\footnotemark[2]} &- &- &- &0.97 &0.97 &0.97 &0.98 &0.97 &0.97 &0.98 &0.97 &0.97 \\
\textbf{Malayalam (ml)} &0.85 &0.86 &0.84 &0.9 &0.92 &0.91 &0.92 &0.9 &0.9 &0.92 &0.92 &0.92 \\
\textbf{Punjabi (pa)} &0.94 &0.96 &0.92 &0.96 &0.96 &0.96 &0.97 &0.96 &0.96 &0.96 &0.95 &0.96 \\
\bottomrule
\end{tabular}}
\end{table}

\begin{figure}[]
\centering
\includegraphics[width=\columnwidth]{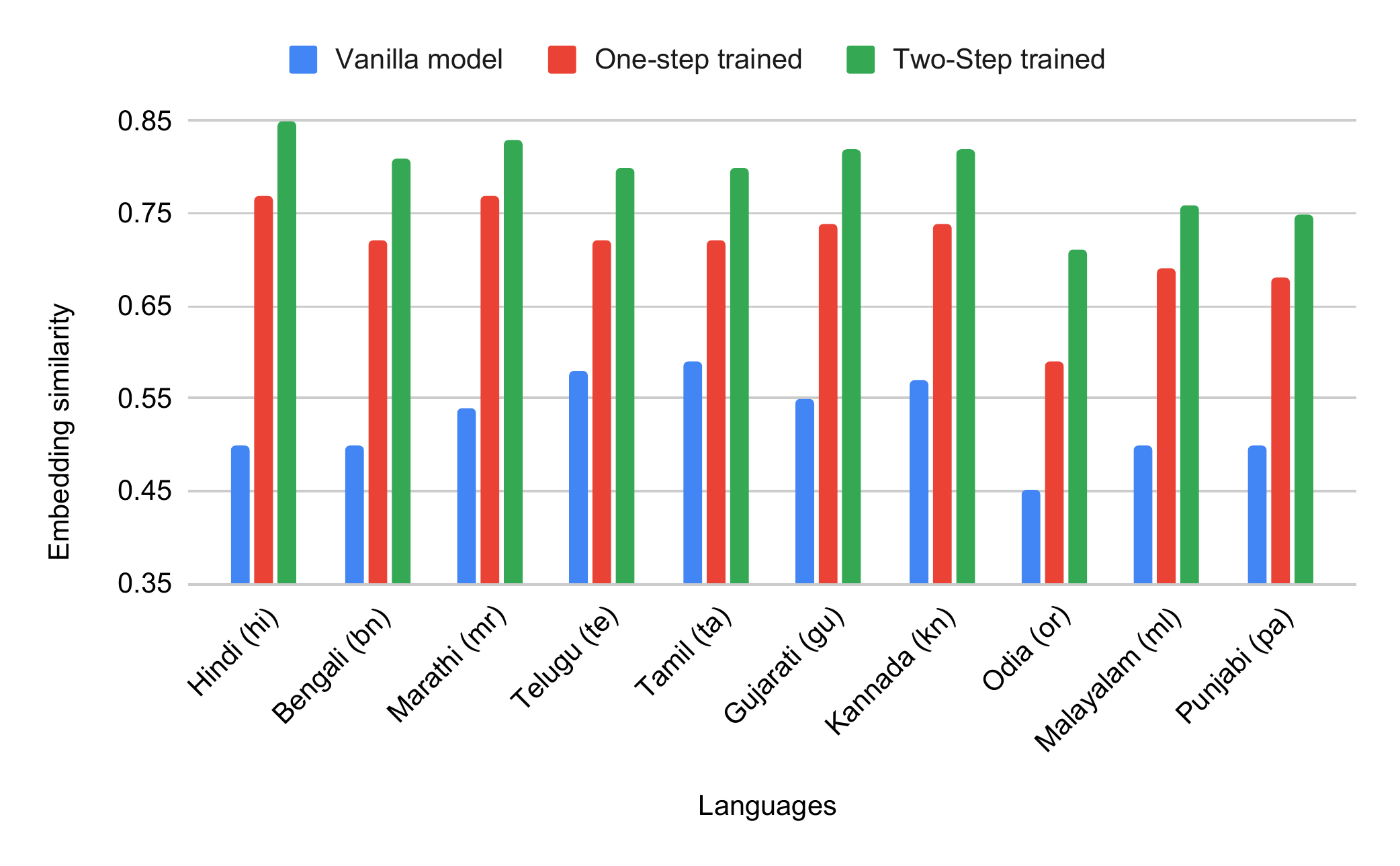}
\caption{Embedding similarity score comparison of SBERT models having monolingual BERT base}\label{fig:3}
\end{figure}

\begin{table}[!htp]\centering
\caption{IndicSBERT: Embedding similarity and classification accuracy results}\label{tab:3}
\scriptsize
\resizebox{\columnwidth}{!}{
\begin{tabular}{|l|c|c|c|c|}\toprule
&\multicolumn{2}{c|}{\textbf{Embedding Similarity}} &\multicolumn{2}{c|}{\textbf{Classification Accuracy}} \\\midrule
\textbf{} &\textbf{IndicSBERT} &\textbf{IndicSBERT-STS} &\textbf{IndicSBERT} &\textbf{IndicSBERT-STS} \\\midrule
\textbf{Hindi (hi)} &0.76 &0.8 &0.68 &0.65 \\
\textbf{Bengali (bn)} & 0.76 & 0.81 & 0.98 & 0.97 \\
\textbf{Marathi (mr)} &0.75 &0.8 &0.98 &0.98 \\
\textbf{Telugu (te)} &0.74 &0.8 &0.99 &0.98 \\
\textbf{Tamil (ta)} &0.74 &0.8 &0.96 &0.95 \\
\textbf{Gujarati (gu)} &0.76 &0.81 &0.99 &0.99 \\
\textbf{Kannada (kn)} &0.76 &0.81 &0.96 &0.95 \\
\textbf{Odia (or)} &0.66 &0.73 &0.97 &0.95 \\
\textbf{Malayalam (ml)} &0.7 &0.76 &0.91 &0.89 \\
\textbf{Punjabi (pa)} &0.7 &0.76 &0.95 &0.96\\
\bottomrule
\end{tabular}}
\end{table}

\begin{figure}[]
\centering
\includegraphics[width=\columnwidth]{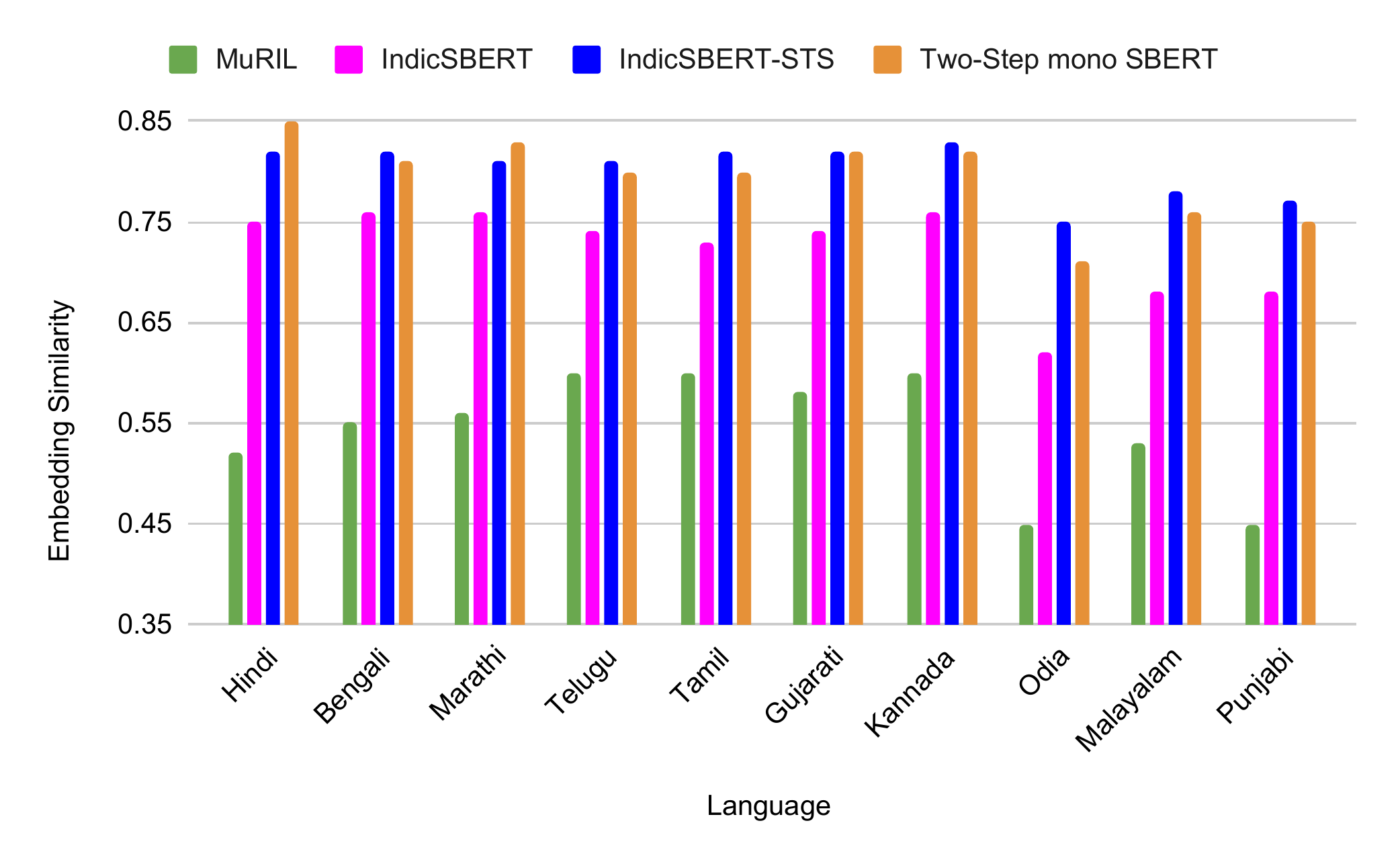}
\caption{Embedding similarity score comparison of MuRIL, IndicSBERT and monolingual SBERT models}\label{fig:4}
\end{figure}

\begin{table}[!htp]\centering
\caption{Zero-shot performance of multilingual models}\label{tab:4}
\scriptsize
\resizebox{\columnwidth}{!}{
\begin{tabular}{|l|c|c|c|c|c|c|c|}\toprule
&\textbf{mBERT} &\textbf{MuRIL} &\textbf{LASER} &\textbf{mpnet-base} &\textbf{LaBSE} &\textbf{IndicSBERT} &\textbf{IndicSBERT-STS} \\\midrule
\textbf{Hindi} &0.49 &0.52 &0.64 &0.79 &0.72 &0.75 &0.82 \\
\textbf{Bengali} &0.5 &0.55 &0.68 &0.66 &0.71 &0.76 &0.82 \\
\textbf{Marathi} &0.47 &0.56 &0.6 &0.75 &0.7 &0.76 &0.81 \\
\textbf{Telugu} &0.53 &0.6 &0.59 &0.64 &0.73 &0.74 &0.81 \\
\textbf{Tamil} &0.49 &0.6 &0.49 &0.65 &0.72 &0.73 &0.82 \\
\textbf{Gujarati} &0.47 &0.58 &0.14 &0.73 &0.73 &0.74 &0.82 \\
\textbf{Kannada} &0.52 &0.6 &0.17 &0.65 &0.72 &0.76 &0.83 \\
\textbf{Odia\footnotemark[2]} &- &0.45 &0.29 &0.48 &0.6 &0.62 &0.75 \\
\textbf{Malayalam} &0.46 &0.53 &0.6 &0.6 &0.66 &0.68 &0.78 \\
\textbf{Punjabi} &0.43 &0.45 &0.12 &0.56 &0.64 &0.68 &0.77 \\
\bottomrule
\end{tabular}}
\end{table}

\footnotetext[2]{ Odia language is not supported by mBERT}
\footnotetext[8]{ Training steps= 0 indicates the vanilla base model, 1 denotes single-step NLI training over the base model, while 2 denotes the two-step trained model}

\begin{figure}[]
\centering
\includegraphics[width=\columnwidth]{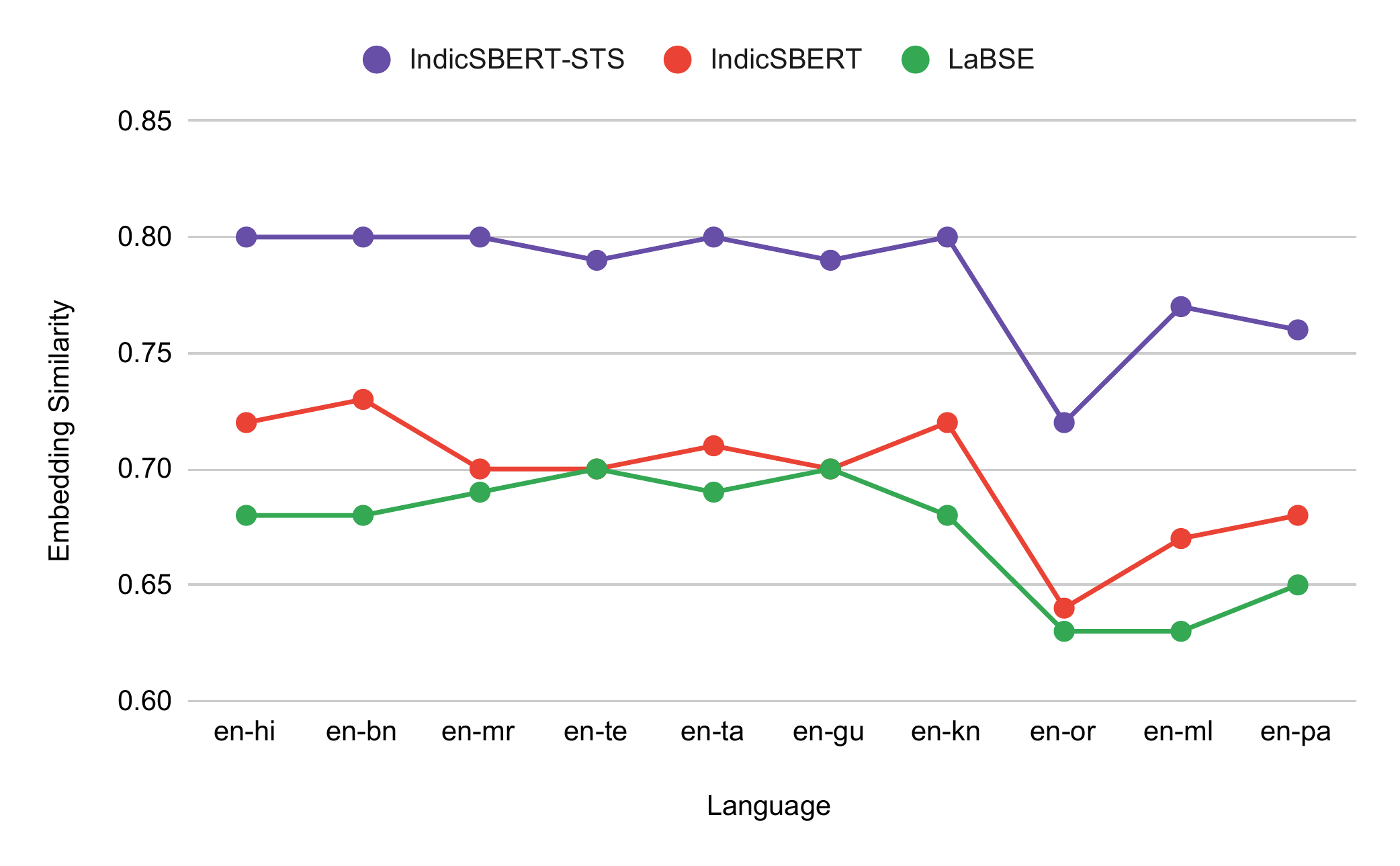}
\caption{Cross-lingual performance of models for English with Indian languages}\label{fig:6}
\end{figure}

\begin{figure*}[]\centering
\centering
\includegraphics[scale=0.5]{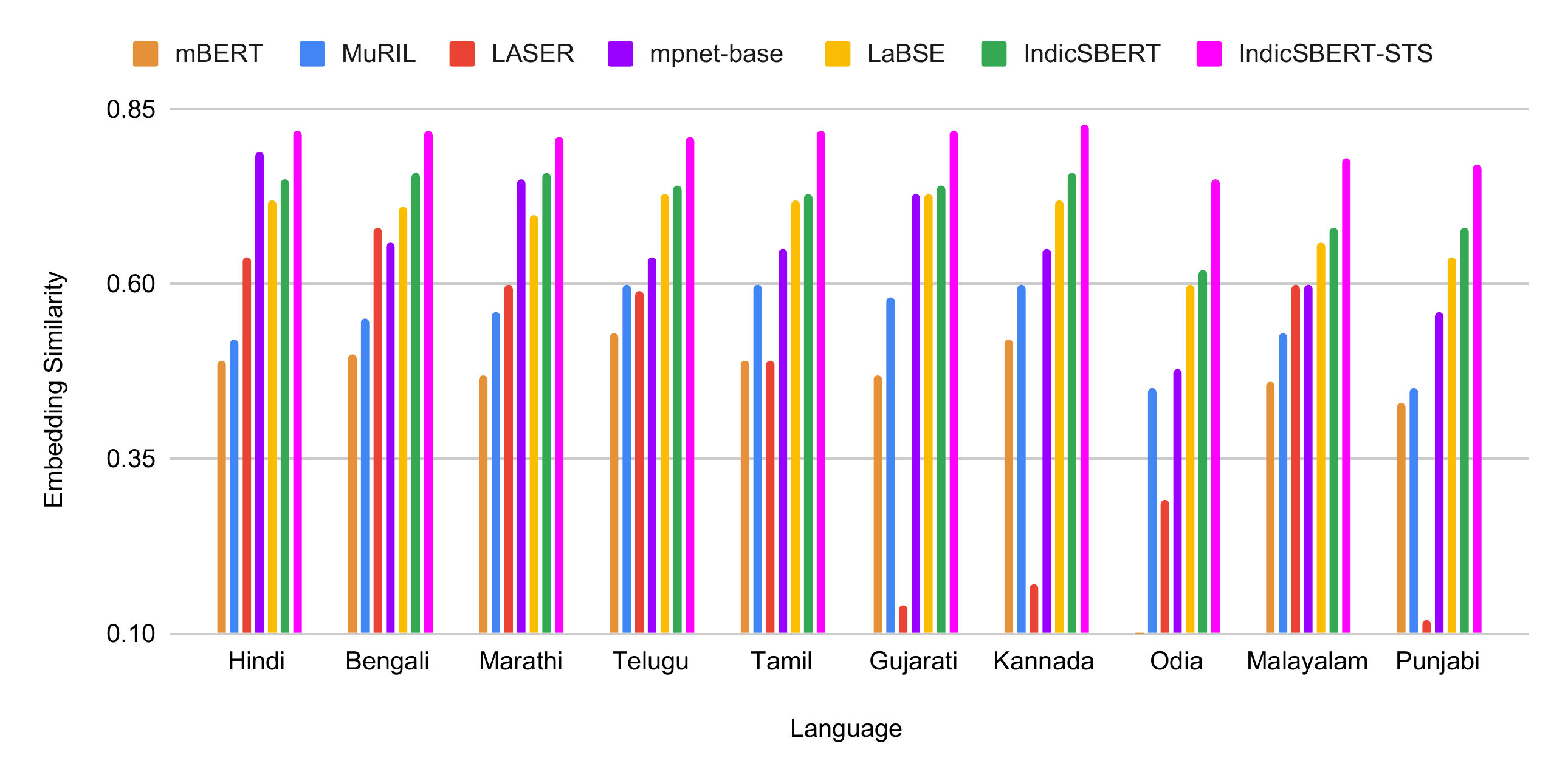}
\caption{Embedding Similarity score comparison of multilingual models}\label{fig:5}
\end{figure*}

\section{Evaluation}
\subsection{Evaluation Methodology}
We evaluate the SBERT models on the basis of the Embedding Similarity score as well as classification accuracy. The Embedding Similarity evaluation is performed by calculating the Pearson and Spearman rank correlation of the embeddings for different similarity metrics with the gold-standard scores. A high score in embedding similarity indicates that the embeddings being compared are of high quality in relation to the benchmark embeddings. 

In our experiment, we use the cosine similarity metric and the value of Spearman correlation to evaluate sentence similarity models. The choice of cosine similarity is based on its superiority compared to traditional distance metrics such as Euclidean or Manhattan distance. Unlike these distance metrics, cosine similarity measures the cosine of the angle between the vectors representing the sentences and considers only their direction, making it less affected by scaling and more computationally efficient. Additionally, cosine similarity takes into account shared terms and contexts, providing a more accurate representation of semantic relationships between sentences. Spearman correlation, on the other hand, is used in preference to Pearson correlation because it is more robust to non-linear relationships and handles ties in data. Unlike Pearson correlation, which assumes a linear relationship, Spearman correlation measures the rank relationship between two variables, making it better equipped to accurately assess a model's performance in cases where the relationship is non-linear.

In this study, the text classification datasets were used to evaluate the performance of BERT and SBERT models in generating sentence embeddings. The K Nearest Neighbors (KNN) algorithm was applied to classify the texts based on proximity. The Minkowski distance, a generalized form of both the Euclidean and Manhattan distances, is employed. The optimal value of k was determined using a validation dataset and then used to calculate the accuracy of the test dataset, with results reported in Tables \ref{tab:2}, \ref{tab:3}.

\subsection{Evaluation Results \& Discussion}
Table \ref{tab:1} presents the Embedding Similarity scores of monolingual SBERT models, while the classification accuracies are displayed in Table \ref{tab:2}. Table \ref{tab:3} presents the similarity and accuracy results of the IndicSBERT. Table \ref{tab:4} compares the zero-shot performance of various multilingual models with that of IndicSBERT, while the superior cross-lingual performance of IndicSBERT is shown in Table \ref{tab:5}. Our observations from these tables are discussed below.
\begin{itemize}[leftmargin=0pt]
\item[] 1. \textbf{AVG pooling shows better performance than CLS}\\We find that monolingual SBERT models generate superior embedding similarity scores when AVG pooling is utilized instead of CLS, across all 10 Indic languages. The same trend is observed for IndicSBERT, where AVG pooling produces better results than CLS for embedding similarity scores. Hence, the AVG pooling values are reported in this work.\\

\item[] 2. \textbf{NLI + STS training works better}\\
Fine-tuning the pre-trained models using NLI followed by STSb gives an upper hand over single-step training using the NLI dataset alone. Figure \ref{fig:3} compares the embedding similarities for the Vanilla, One-step trained, and Two-step trained monolingual models. We observe that the Two-step trained models surpass the one-step and Vanilla models in terms of performance across all 10 Indic languages. Fine-tuning with the STSb dataset results in a significant increase in embedding similarity for the monolingual SBERT models as well as for IndicSBERT, as demonstrated by Tables \ref{tab:1}, \ref{tab:3}, and Figures \ref{fig:3},\ref{fig:4}. Figure \ref{fig:4} demonstrates that the two-step training on IndicSBERT, which employs MuRIL as its base model, increases the embedding similarity scores nearly two-fold in comparison to the vanilla MuRIL model. While we mainly focus on cross-lingual performance in this work, similar observations in the context of monolingual SBERT have been thoroughly documented in \cite{joshi2022l3cubemahasbert}.\\

\item[] 3. \textbf{SBERT models trained on synthetic corpus work well with real-world classification datasets}\\We evaluate our sentence-BERT models on real-world news classification datasets to ensure that the models do not overfit the noise in the synthetic corpus. The results presented in Tables \ref{tab:2} and \ref{tab:3} indicate that SBERT models trained on translated corpora perform competitively compared to their original base models on classification datasets. The classification accuracy is neither improved nor deteriorated due to the two-step training.\\

\item[] 4. \textbf{Multilingual Indic-SBERT is competitive with monolingual SBERT models}\\Our experiments indicate that the multilingual IndicSBERT model demonstrates equivalent or better performance compared to monolingual SBERT models in terms of embedding similarity scores, as evidenced by Tables \ref{tab:1}, \ref{tab:3}. In \ref{fig:4}, we observe that both the IndicSBERT-STS and two-step monolingual SBERT models perform comparably, with slight performance differences for certain languages. Except for Hindi, Marathi and Gujarati languages, the IndicSBERT-STS outperforms the SBERT models of the respective languages. This shows that the languages have assisting capabilities and the gains are higher for extremely low-resource languages like Odia and Punjabi.\\

\item[] 5. \textbf{IndicSBERT works significantly better than existing multilingual models}\\ Figure \ref{fig:5}, as well as Table \ref{tab:4}, compare the zero-shot embedding similarities of mBERT, MuRIL, LASER, multilingual-mpnet-base, LaBSE, and IndicSBERT models on STSb for all 10 Indic languages, with the IndicSBERT based models clearly outperforming the others. Both IndicSBERT and IndicSBERT-STS produce richer embeddings than the publicly available LaBSE, which is shown in Table ~\ref{tab:4}. Thus, the IndicSBERT is the best-performing model among all the other publicly available multilingual models despite having the lowest number of trainable parameters.\\

\item[] 6. \textbf{IndicSBERT shows exceptional cross-lingual properties, outperforming the LaBSE}\\The results presented in the Table ~\ref{tab:5} and Figure \ref{fig:6} demonstrate IndicSBERT's robust cross-lingual performance across all language pairs, surpassing the performance of LaBSE by a significant margin. Overall, the multilingual IndicSBERT model demonstrates proficiency in processing both monolingual and multilingual datasets. This versatility enables the development of language-independent NLP applications that can seamlessly work across multiple Indian languages. In addition, IndicSBERT has the potential to enhance the precision and effectiveness of cross-lingual information retrieval systems and semantic search engines as it can handle queries and documents in multiple Indian languages. This characteristic holds particular importance for countries such as India, where multilingual communication is common, and organizations face the challenge of accommodating diverse language requirements.\\

\item[] 7. \textbf{Multilingual models are indeed cross-lingual learners, the enhancement of cross-lingual properties is generalizable to non-Indic languages}
\\The performance of mBERT with mixed language NLI training on diverse languages like English, Hindi, German, and French is presented in Table ~\ref{tab:6}. The results demonstrate a considerable improvement in the cross-lingual performance of the one-step trained model as compared to the vanilla mBERT. These findings support the effectiveness of the proposed mixed-language training technique in producing models with enhanced cross-lingual properties not only for Indic languages but also for other languages. 

\end{itemize}

\begin{table}[!htp]\centering
\caption{Cross-lingual performance of mBERT, single-step trained for 4 languages: Hindi, English, German and French. For every language-pair, the
values reported from top to bottom correspond to One-step mBERT, and vanilla mBERT respectively}\label{tab:6}
\scriptsize
\begin{tabular}{|c|c|c|c|c|}\toprule
&\textbf{Hindi} &\textbf{English} &\textbf{German} &\textbf{French} \\\midrule
\multirow{2}{*}{\textbf{Hindi}} &\textbf{0.68} &\textbf{0.5} &\textbf{0.5} &\textbf{0.48} \\
&0.48 &0.3 &0.3 &0.32 \\\midrule
\multirow{2}{*}{\textbf{English}} &\textbf{0.51} &\textbf{0.77} &\textbf{0.6} &\textbf{0.63} \\
&0.31 &0.5 &0.4 &0.41 \\\midrule
\multirow{2}{*}{\textbf{German}} &\textbf{0.49} &\textbf{0.6} &\textbf{0.7} &\textbf{0.56} \\
&0.3 &0.4 &0.48 &0.39 \\\midrule
\multirow{2}{*}{\textbf{French}} &\textbf{0.49} &\textbf{0.63} &\textbf{0.57} &\textbf{0.72} \\
&0.29 &0.39 &0.37 &0.49 \\
\bottomrule
\end{tabular}
\end{table}

\begin{table*}[!htp]\centering
\caption{Cross-lingual performance of IndicSBERT-STS, IndicSBERT and LaBSE. For every language-pair, the values reported from top to bottom correspond to IndicSBERT-STS, IndicSBERT and LaBSE respectively}\label{tab:5}
\scriptsize
\begin{tabular}{|c|c|c|c|c|c|c|c|c|c|c|c|}\toprule
&English &Hindi &Bengali &Marathi &Telugu &Tamil &Gujarati &Kannada &Oriya &Malayalam &Punjabi \\\midrule
\multirow{3}{*}{English} &\textbf{0.85} &\textbf{0.8} &\textbf{0.8} &\textbf{0.8} &\textbf{0.79} &\textbf{0.8} &\textbf{0.79} &\textbf{0.8} &\textbf{0.72} &\textbf{0.77} &\textbf{0.76} \\
&0.8 &0.72 &0.73 &0.7 &0.7 &0.71 &0.7 &0.72 &0.64 &0.67 &0.68 \\
&0.72 &0.68 &0.68 &0.69 &0.7 &0.69 &0.7 &0.68 &0.63 &0.63 &0.65 \\\midrule
\multirow{3}{*}{Hindi} &\textbf{0.82} &\textbf{0.82} &\textbf{0.79} &\textbf{0.79} &\textbf{0.77} &\textbf{0.78} &\textbf{0.79} &\textbf{0.79} &\textbf{0.72} &\textbf{0.76} &\textbf{0.76} \\
&0.72 &0.75 &0.71 &0.7 &0.68 &0.68 &0.69 &0.69 &0.62 &0.65 &0.68 \\
&0.7 &0.72 &0.69 &0.7 &0.7 &0.69 &0.71 &0.68 &0.62 &0.62 &0.64 \\\midrule
\multirow{3}{*}{Bengali} &\textbf{0.82} &\textbf{0.79} &\textbf{0.82} &\textbf{0.79} &\textbf{0.77} &\textbf{0.77} &\textbf{0.79} &\textbf{0.79} &\textbf{0.73} &\textbf{0.76} &\textbf{0.76} \\
&0.73 &0.7 &0.76 &0.7 &0.68 &0.68 &0.7 &0.7 &0.63 &0.65 &0.67 \\
&0.69 &0.69 &0.71 &0.69 &0.7 &0.69 &0.71 &0.69 &0.64 &0.64 &0.66 \\\midrule
\multirow{3}{*}{Marathi} &\textbf{0.8} &\textbf{0.78} &\textbf{0.78} &\textbf{0.81} &\textbf{0.76} &\textbf{0.77} &\textbf{0.78} &\textbf{0.78} &\textbf{0.72} &\textbf{0.75} &\textbf{0.75} \\
&0.7 &0.7 &0.7 &0.76 &0.67 &0.66 &0.69 &0.69 &0.62 &0.65 &0.67 \\
&0.68 &0.68 &0.69 &0.7 &0.69 &0.68 &0.7 &0.68 &0.63 &0.64 &0.65 \\\midrule
\multirow{3}{*}{Telugu} &\textbf{0.79} &\textbf{0.77} &\textbf{0.77} &\textbf{0.76} &\textbf{0.81} &\textbf{0.77} &\textbf{0.76} &\textbf{0.78} &\textbf{0.71} &\textbf{0.74} &\textbf{0.73} \\
&0.72 &0.68 &0.68 &0.68 &0.74 &0.68 &0.67 &0.69 &0.6 &0.64 &0.65 \\
&0.7 &0.7 &0.7 &0.7 &0.73 &0.7 &0.71 &0.69 &0.63 &0.64 &0.66 \\\midrule
\multirow{3}{*}{Tamil} &\textbf{0.8} &\textbf{0.77} &\textbf{0.77} &\textbf{0.76} &\textbf{0.76} &\textbf{0.82} &\textbf{0.76} &\textbf{0.77} &\textbf{0.7} &\textbf{0.75} &\textbf{0.73} \\
&0.71 &0.67 &0.67 &0.67 &0.67 &0.73 &0.65 &0.68 &0.58 &0.64 &0.64 \\
&0.69 &0.7 &0.69 &0.69 &0.7 &0.72 &0.7 &0.68 &0.62 &0.62 &0.64 \\\midrule
\multirow{3}{*}{Gujarati} &\textbf{0.8} &\textbf{0.79} &\textbf{0.78} &\textbf{0.79} &\textbf{0.76} &\textbf{0.76} &\textbf{0.82} &\textbf{0.77} &\textbf{0.73} &\textbf{0.74} &\textbf{0.76} \\
&0.7 &0.69 &0.69 &0.69 &0.67 &0.66 &0.74 &0.68 &0.6 &0.63 &0.67 \\
&0.7 &0.7 &0.7 &0.69 &0.7 &0.69 &0.73 &0.68 &0.63 &0.63 &0.66 \\\midrule
\multirow{3}{*}{Kannada} &\textbf{0.8} &\textbf{0.77} &\textbf{0.77} &\textbf{0.77} &\textbf{0.77} &\textbf{0.77} &\textbf{0.76} &\textbf{0.83} &\textbf{0.7} &\textbf{0.75} &\textbf{0.73} \\
&0.71 &0.68 &0.69 &0.68 &0.68 &0.67 &0.66 &0.76 &0.59 &0.65 &0.64 \\
&0.68 &0.67 &0.68 &0.68 &0.69 &0.67 &0.69 &0.72 &0.62 &0.62 &0.64 \\\midrule
\multirow{3}{*}{Oriya} &\textbf{0.72} &\textbf{0.71} &\textbf{0.72} &\textbf{0.7} &\textbf{0.7} &\textbf{0.7} &\textbf{0.72} &\textbf{0.7} &\textbf{0.75} &\textbf{0.68} &\textbf{0.7} \\
&0.62 &0.61 &0.61 &0.6 &0.6 &0.58 &0.6 &0.6 &0.62 &0.56 &0.6 \\
&0.6 &0.59 &0.6 &0.6 &0.6 &0.6 &0.61 &0.6 &0.6 &0.58 &0.59 \\\midrule
\multirow{3}{*}{Malayalam} &\textbf{0.77} &\textbf{0.74} &\textbf{0.75} &\textbf{0.74} &\textbf{0.74} &\textbf{0.75} &\textbf{0.73} &\textbf{0.74} &\textbf{0.69} &\textbf{0.78} &\textbf{0.7} \\
&0.68 &0.65 &0.66 &0.66 &0.65 &0.65 &0.63 &0.65 &0.57 &0.68 &0.62 \\
&0.64 &0.62 &0.64 &0.64 &0.65 &0.64 &0.65 &0.64 &0.6 &0.66 &0.6 \\\midrule
\multirow{3}{*}{Punjabi} &\textbf{0.76} &\textbf{0.76} &\textbf{0.76} &\textbf{0.75} &\textbf{0.73} &\textbf{0.74} &\textbf{0.76} &\textbf{0.74} &\textbf{0.7} &\textbf{0.71} &\textbf{0.77} \\
&0.68 &0.67 &0.67 &0.66 &0.65 &0.64 &0.66 &0.66 &0.6 &0.61 &0.68 \\
&0.65 &0.63 &0.66 &0.65 &0.66 &0.65 &0.66 &0.64 &0.62 &0.6 &0.64 \\
\bottomrule
\end{tabular}
\end{table*}

\section{Conclusion}
Our research addresses the crucial gap in the availability of high-quality language models for low-resource Indian languages. We have presented a range of SBERT models for ten popular Indian languages, trained using synthetic corpus. They have been evaluated based on their embedding similarity with the translated standard STSb dataset and accuracies over text classification datasets. Our results demonstrate that the monolingual SBERT models outperform vanilla BERT models in terms of embedding similarity. Additionally, we have developed the multilingual IndicSBERT, which exhibits strong cross-lingual performance and outperforms existing multilingual models such as LaBSE and paraphrase-multilingual-mpnet-base-v2. This is a significant contribution to the field of IndicNLP, particularly in the context of the world becoming more globalized, and the need for accurate and efficient multilingual NLP models. While doing so we present a simple and clean approach to train cross-lingual sentence BERT models using only translated monolingual datasets and vanilla multilingual BERT.

Indian languages pose a unique challenge, being diverse and having low-resource corpora. Our study highlights the effectiveness of the two-step training method in developing both monolingual SBERT models and the multilingual IndicSBERT. Its robust cross-lingual capability makes IndicSBERT a superior choice for applications that require accurate and efficient multilingual NLP.

As part of this publication, we are releasing the monolingual SBERTs and the multilingual IndicSBERT, which will open up new possibilities for NLP research and applications in low-resource Indian languages. In summary, our research contributes to the development of high-quality language models for Indian languages and highlights the importance of combining the power of sentence-level embeddings with the ability to handle multiple languages to achieve optimal results in multilingual NLP applications.

\begin{acks}
This work was done under the L3Cube Pune mentorship
program. We would like to express our gratitude towards
our mentors at L3Cube for their continuous support and
encouragement. This work is a part of the MahaNLP project \cite{joshi2022l3cubemahanlp}.
\end{acks}

\bibliographystyle{ACM-Reference-Format}
\bibliography{main}


\end{document}